\documentclass[11pt,a4paper]{article}

\usepackage{graphicx}    
\usepackage{amsmath}  
\usepackage{geometry}
\usepackage{hyperref}

\usepackage[T1]{fontenc}
\usepackage[latin1]{inputenc} 
\usepackage{xcolor}
\usepackage{amssymb}
\usepackage{bm}

\usepackage{cite}
\usepackage{multirow}

\usepackage{fancyhdr}
\def\etal{\emph{et al.}}

\title{ \textbf{EPI-based Oriented Relation Networks for Light Field Depth Estimation}}

\author{Kunyuan Li \\ \scriptsize{2018110972@mail.hfut.edu.cn} \and Jun Zhang\thanks{Corresponding author} \\ \scriptsize{zhangjun@hfut.edu.cn} \and Rui Sun \\ \scriptsize{sunrui@hfut.edu.cn} \and Xudong Zhang \\ \scriptsize{xudong@hfut.edu.cn} \and Jun Gao \\ \scriptsize{gaojun@hfut.edu.cn}}

\date{ 
     }

\begin{document}

\maketitle

\begin{abstract}
Light field cameras record not only the spatial information of observed scenes but also the directions of all incoming light rays.
The spatial and angular information implicitly contain geometrical characteristics such as multi-view or epipolar geometry, which can be exploited to improve the performance of depth estimation. An Epipolar Plane Image (EPI), the unique 2D spatial-angular slice of the light field, contains patterns of oriented lines. The slope of these lines is associated with the disparity. Benefiting from this property of EPIs, some representative methods estimate depth maps by analyzing the disparity of each line in EPIs. 
However, these methods often extract the optimal slope of the lines from EPIs while ignoring the relationship between neighboring pixels, which leads to inaccurate depth map predictions. Based on the observation that an oriented line and its neighboring pixels in an EPI share a similar linear structure, we propose an end-to-end fully convolutional network (FCN) to estimate the depth value of the intersection point on the horizontal and vertical EPIs. 
Specifically, we present a new feature-extraction module, called {\bf Oriented Relation Module (ORM)}, that constructs the relationship between the line orientations. 
To facilitate training, we also propose a refocusing-based data augmentation method to obtain different slopes from EPIs of the same scene point. Extensive experiments verify the efficacy of learning relations and show that our approach is competitive to other state-of-the-art methods. The code and the trained models are available at \url{https://github.com/lkyahpu/EPI_ORM.git}.

\end{abstract}

\section{Introduction}
\label{sec:intro}
Light field cameras record both 2D spatial and 2D angular information of the observed scene~\cite{LFrendering}.
The lenslet-based light field camera~\cite{Ren2006Digital}, a compact and hand-held light field camera, is able to achieve the dense sampling of the viewpoints by utilizing a micro-lens array inserted between the main lens and the photo sensor.
The captured 4D light field data implicitly contains geometrical characteristics such as multi-view geometry or epipolar geometry, which
has attracted much attention in recent years to improve the performance of depth estimation from light fields.
\begin{figure}[t!]
\setlength{\abovecaptionskip}{-0.5cm}
\setlength{\belowcaptionskip}{-0.3cm}
	\begin{center}
		\includegraphics[width=1\linewidth]{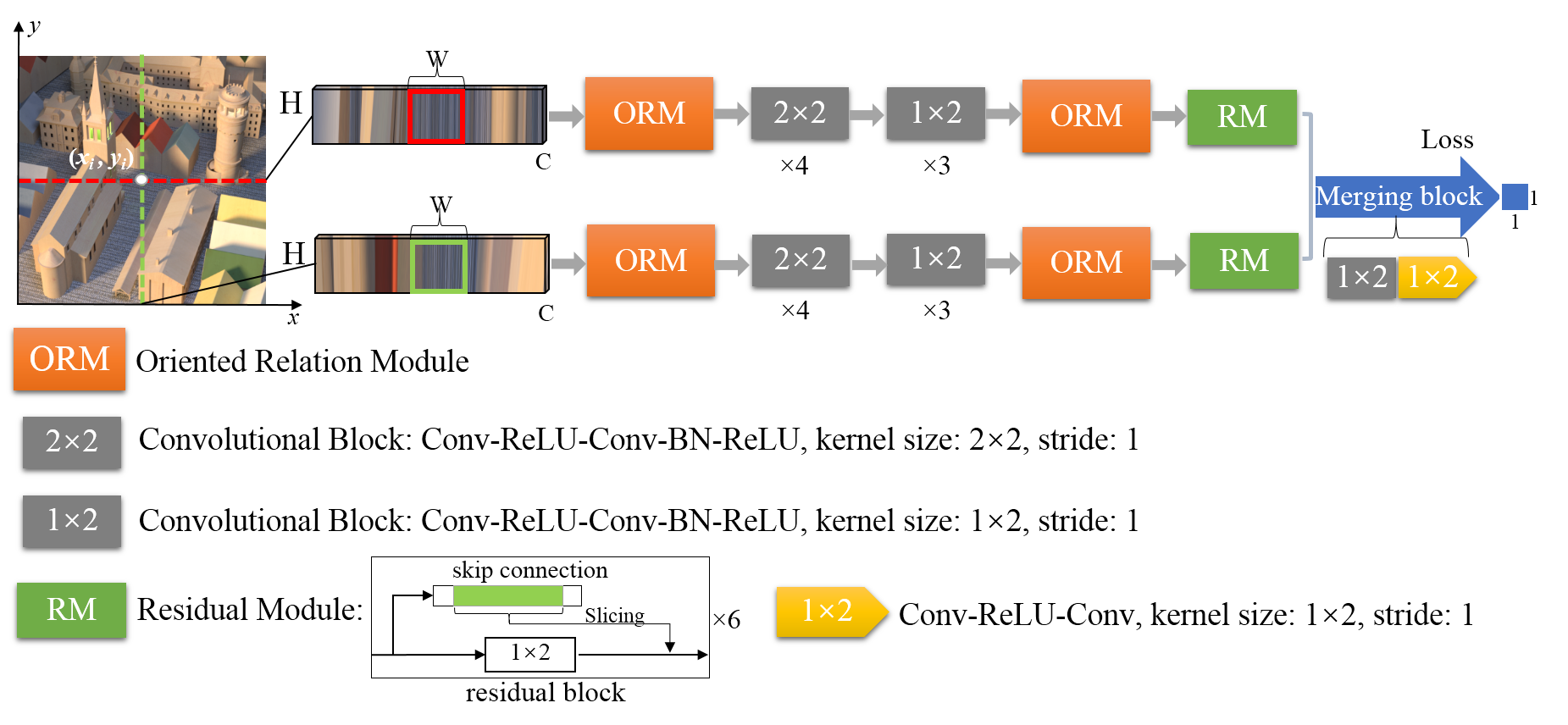}
	\end{center}
	\caption{An overview of the proposed network architecture. The input is a pair of EPI patches obtained from the horizontal and vertical EPIs. Each branch consists of two oriented relation modules (ORMs), seven convolutional blocks, and a residual module (RM). The output of the two branches is integrated by a merging block to estimate the depth value of each pixel.}
	\label{fig:network}
\end{figure}

To visualize light fields and extract light field features, the 4D light field data is often converted into various 2D images such as multi-view sub-aperture images~\cite{Hanrahan2005Light}, Epipolar Plane Images (EPIs)~\cite{LFrendering}, and focal stacks~\cite{focalStack2015}.
Some representative methods \cite{Tao2013,Wang2015,Williem2016} exploit different depth cues from sub-aperture images and focal stacks for depth estimation.
However, it is difficult to acquire dense and accurate depth maps from the lenslet-based cameras owing to the optical distortions~\cite{op2013} and the narrow baseline~\cite{Ren2006Digital} between sub-aperture images. 
Besides, these methods are usually accompanied by heavy computational burdens and carefully-designed optimization measures. 
To avoid these issues, some methods~\cite{Wanner2014,ZhangSPO,HeberandPock2016,Luo} exploit EPIs that exhibit patterns of oriented lines with constant colors to visualize light fields.
Each of these lines corresponds to the projection of a single 3D scene point, and its slope is called disparity \cite{Wanner2012Globally}. 
Therefore, one can infer the depth of the corresponding scene point by analyzing the disparity of the oriented line in the EPI. 
Moreover, the oriented line and its neighboring pixels share the similar linear structure, which is beneficial to estimate the slope of the EPI by constructing the relationship between the center region in the EPI and its neighborhood. 
Nonetheless, current methods predict depth maps by extracting the optimal slope of EPIs while ignoring the relationship between neighboring pixels in EPIs, which makes the results inaccurate. 
It has been well recognized that the relation information is capable of offering important visual cues for computer vision tasks, such as spatial and channel relations in semantic segmentation~\cite{Mou2019} and object detection~\cite{Hu2018}, and temporal relations in activity recognition~\cite{Zhou2018}. 

In this paper, we propose an end-to-end fully convolutional network to estimate the depth value of the intersection point on the horizontal and vertical EPIs, as shown in Figure \ref{fig:network}.
We design a Siamese network without sharing weights (i.\ e. pseudo-Siamese~\cite{siamese2015}) so that the convolution weights of the horizontal and vertical EPIs can be learned separately. 
Specifically, we propose a new feature extraction module, called {\bf Oriented Relation Module (ORM)}, to learn and reason about the relationship between oriented lines in EPIs by extracting oriented relation features between the center pixel and its neighborhood from EPI patches.
The proposed method can be considered as the first work on modeling relation features in EPIs, which is novel and different from existing relation models in two aspects: First, existing works~\cite{Mou2019,Zhou2018} focus on modeling temporal relation between frames and spatial relations between pixels. In contrast, our method proposes the geometric relation between line orientations in EPI patches, which is beneficial to extract the accurate slopes of EPIs for light field depth estimation. Second, the proposed method models dependencies between oriented lines, without making any assumptions on their feature distributions and locations.
Our network is trained using the 4D light field benchmark dataset~\cite{HCIdataset2017}, where the ground truth disparities are available.
However, we find that it is hard to train such a deep network with insufficient data. 
To mitigate this issue, we propose a data augmentation method by refocusing EPIs so that EPIs with different slopes as well as the corresponding ground truth disparities can be obtained at the same scene point.
We show that the newly proposed ORM and EPI-based data augmentation can bring performance boost for light field depth estimation. 

\section{Related Work}
Conventional depth estimation from light fields mainly relies on different assumptions \cite{Wang2015,focalStack2015} and handcrafted depth features ~\cite{Tao2013,Williem2016} based on sub-aperture images and focal stacks. 
In this section, we restrict ourselves to methods that exploit EPIs, and review some representative works with relation reasoning.

\vspace{-5 pt}
\paragraph{Light field depth estimation based on EPIs.}
There exist a few methods that exploit the EPI for light field depth estimation due to its linear structure associated with depth~\cite{Wanner2012Globally}.
For example, Wanner \etal~\cite{Wanner2014} used a structured tensor to compute the slope of each line in vertical and horizontal EPIs. Zhang \etal~\cite{ZhangSPO} introduced the Spinning Parallelogram Operator (SPO) to find matching lines in EPIs. The lines with different slopes are located by maximizing the distribution distances of the regions. 
Zhang \etal~\cite{YZhang} located the optimal slope of each line segmentation on EPIs by using the locally linear embedding.
Differing from these methods, some methods applied CNNs to extract light field features from EPIs.
Sun \etal \cite{Sun2016} presented a data-driven approach to estimate the object depths from an enhanced EPI feature using CNN. 
Heber and Pock \cite{HeberandPock2016} used CNNs for predicting 2D per-pixel hyperplane slope orientations in EPIs. 
Based on this work, Heber \etal \cite{Heber2016,Heber2017} improved their work by utilizing an U-shaped network and EPI volumes to predict the depth map. 
Luo \etal \cite{Luo} designed an EPI-patch based CNN architecture to estimate the depth of each pixel. 
Feng \etal~\cite{Feng2018} proposed a two-stream network that learns to estimate the depth values of multiple correlated neighborhood pixels from EPI patches.
Shin \etal~\cite{EPINET} introduced a multi-stream network to extract features for epipolar property of four viewpoints with horizontal, vertical and both diagonal directions. 
This method reaches state-of-the-art results on the 4D light field benchmark~\cite{HCIdataset2017}. One of the most recent works by Leistner \etal~\cite{EPI-shift} shift the light field stack to retain a small receptive field, which improves the performance of depth estimation for large-disparity light fields.
Some of these previous methods \cite{Heber2016,Luo,Feng2018,EPI-shift} require data pre-processing and subsequent optimization.
In contrast, we present an end-to-end fully convolutional network architecture to predict the depth values of center pixels from the corresponding horizontal and vertical EPIs. We explore the similar linear structure information in EPIs and model the relationship between the oriented lines and their neighboring pixels, which help to estimate the slope of the oriented line. 

\vspace{-10 pt}
\paragraph{Relation modeling.}
A few recent papers \cite{Zhou2018,Hu2018,Mou2019} have shown that relations have been exploited to improve the performance of computer vision tasks.
Zhou \etal~\cite{Zhou2018} proposed a temporal relation network to learn and reason about temporal dependencies between video frames at multiple time scales. Hu \etal~\cite{Hu2018} proposed an object relation module to model relationships between sets of objects for object detection. Mou \etal~\cite{Mou2019} proposed the spatial and channel relation modules to learn and reason about global relationships between any two spatial positions or feature maps, and then produced relation-augmented feature representations for semantic segmentation.
Motivated by these works, we propose a oriented relation module to construct the relationship between the center pixel and its neighborhood in the EPI, which allows the network to explicitly learn the relationship between the line orientations and improve the performance of depth estimation.

\vspace{-3 pt}
\section{Proposed Method}
In this paper, we present an end-to-end fully convolutional network to predict the depth values of center pixels in EPIs of light fields. Two branches are designed to process the horizontal and vertical EPIs separately. The newly proposed oriented relation module is capable of modeling the relationships between the neighboring pixels in EPIs. A refocusing-based EPI augmentation method is also proposed to facilitate training and improve the performance of depth estimation.
An overview of the network architecture is shown in Figure \ref{fig:network}.

\subsection{EPI Patches for Learning}
\begin{figure}[t!]
\setlength{\abovecaptionskip}{-0.5cm}
\setlength{\belowcaptionskip}{0.3cm}
\begin{center}
\includegraphics[width=1\linewidth]{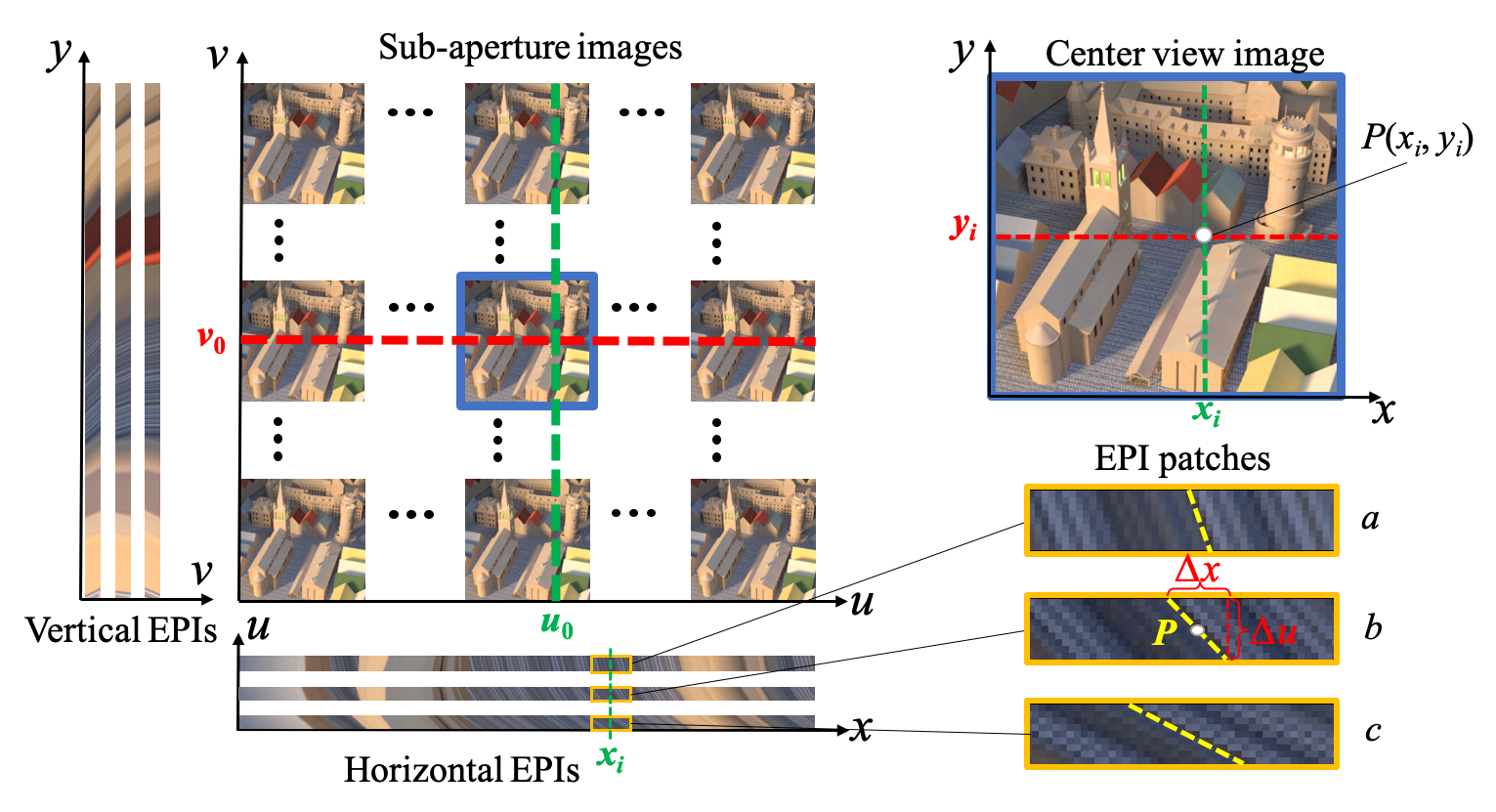}
\end{center}
\caption{EPIs from the light field: each image in angular coordinates $(u,v)$ yields a sub-aperture view of the scene.
Given a pixel $P(x_i,y_i)$ in the spatial coordinate, its horizontal or vertical EPIs are obtained by fixing the view $v_0$ or $u_0$, respectively. 
Three pairs of horizontal and vertical EPIs at different refocused depths are shown. The similar linear structure information between the oriented line marked by the yellow and its neighboring pixels is shown among three EPI patches. The disparity $\frac{\Delta x}{\Delta u}$ of the EPI patch $b$ describes the pixel shift of the scene point $P$ when moving between the views.}
\label{fig:EPI}
\end{figure}

The light field, indicated as $L(u,v,x,y)$, is generally represented by the two-plane parameterization~\cite{LFrendering}. 
Here, $(x,y)$ and $(u,v)$ are spatial and angular coordinates, respectively. 
The central sub-aperture (center view) image is formed by the rays passing through the optical center of the camera main lens ($u = u_0 ,v = v_0$). 
As shown in Figure \ref{fig:EPI}, given a pixel $P(x_i,y_i)$ in the center view image, the horizontal EPI of the row view $v_0$ can be formulated as $L(u,v_0,x,y_i)$, which is centered at $(u_0,x_i)$. 
Similarly, the vertical EPI of the column view $u_0$, with the center at  $(v_0,y_i)$, is written as $L(u_0,v,x_i,y)$.

$Z$ can be obtained by analyzing the slope $\frac{\Delta x}{\Delta u}$ of the line ~\cite{Wanner2014},
\begin{equation}
\Delta x = -\frac{\Delta u}{Z} f
\label{eq:depth}
\end{equation}
where $f$ is the focal distance and $Z$ is the depth value of the point $P$.
The slope of the oriented line is shown in the EPI patch $b$ of Figure \ref{fig:EPI}.

To learn the slope of the oriented line of $P(x_i, y_i)$, we extract patches of size $H \times W \times C$ from $L(u,v_0,x,y_i)$ and $L(u_0,v,x_i,y)$ as inputs.
Here, $H$ and $W$ indicate height and width of the patch, respectively, and $C$ is the channel dimension. The size of the patch is determined by the range of disparities.
The proposed network predicts the depth of the center pixel from the pair of EPI patches.

\subsection{Network Architecture}
As shown in Figure \ref{fig:network}, the proposed network shares the similar structure with the pseudo-Siamese network proposed in ~\cite{siamese2015}, where two branches are designed to learn the weights for the horizontal and vertical EPI patches, respectively. 
Each branch contains two oriented relation modules (ORMs), a set of seven convolutional blocks, a residual module (RM), and a merging block. The ORM will be discussed in Sect.~\ref{sec:orm}.
The convolutional block is composed of `Conv-ReLU-Conv-BN-ReLU'. 
To handle the small EPI slope, we apply the convolutional filters with size of $2 \times 2$ or $1 \times 2$ and stride $1$ to measure a small depth value.
However, detailed information of the EPI slope is lost as the network goes deeper.
Inspired by the residual learning \cite{He2016} that can introduce detailed information of the shallower layer into the deeper layer and effectively improve the network performance, we design a residual module for each branch.
The residual module consists of six residual blocks, each of which consists of one convolutional block and one skip connection. 
We take a slicing operation to implement the skip connection by extracting the center region of the input feature.
The final merging block, containing two different convolutional blocks (`Conv-ReLU-Conv-BN-ReLU' and `Conv-ReLU-Conv'), is used for fusing the horizontal and vertical EPI features to predict the depth value of each pixel. 

\subsection{Oriented Relation Module}
\label{sec:orm}
\begin{figure}[t!]
\setlength{\abovecaptionskip}{-0.5cm}
	\begin{center}
		\includegraphics[width=1\linewidth]{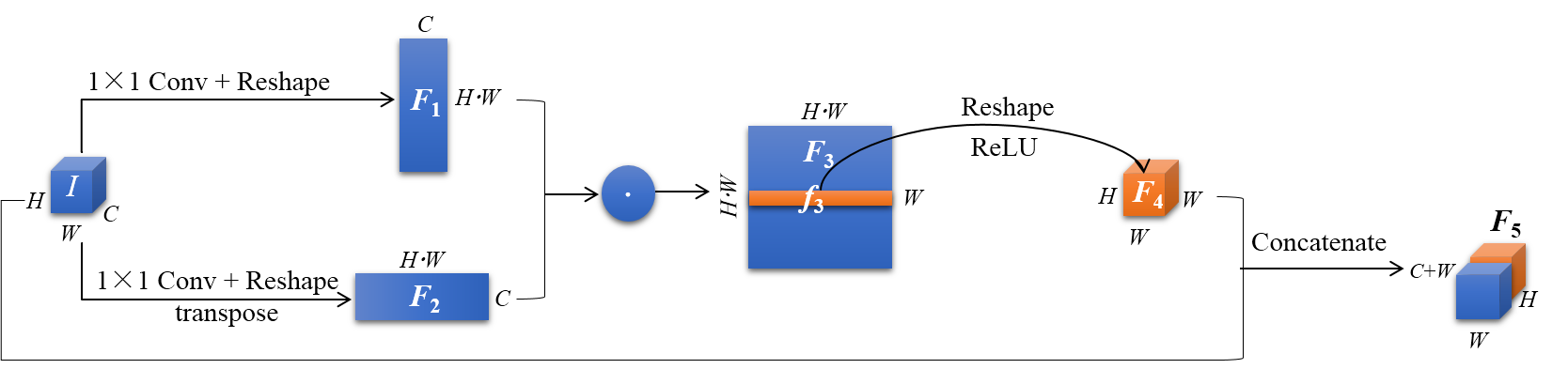}
	\end{center}
	\caption{The proposed oriented relation module.}
	\label{fig:ORM}
\end{figure}

We propose a new Oriented Relation Module (ORM) to reason about the relationship between the center pixel and its neighborhood in each EPI patch.
As shown in Figure \ref{fig:ORM}, given an EPI patch $\bm{I}$ of size $H \times W \times C$, we apply two single-layer convolutions of $1 \times 1$ kernel size to model a compact relationship in the EPI patch.
The output features are converted into $\bm{F_1}$ and $\bm{F_2}$, respectively, which are followed by a dot product to construct the oriented relation feature $\bm{F_3}$ of size $(H\cdot W) \times (H\cdot W)$.
Furthermore, to obtain the relationship between the center pixel and its neighborhood in $\bm{F_3}$, we extract the feature $\bm{f_3}$ of size $W \times (H\cdot W)$ from the relational feature $\bm{F_3}$.
Then, we apply the reshaping and ReLU activation on the feature $\bm{f_3}$ to obtain a new feature $\bm{F_4}$ of size $W \times H \times W$. 
Finally, we concatenate the original EPI patch $\bm{I}$ with the feature $\bm{F_4}$ to obtain the output feature $\bm{F_5}$ of size $W \times H \times (W + C)$.

\subsection{EPI Refocusing-based Data Augmentation}
\begin{figure}[t!]
\setlength{\abovecaptionskip}{-0.5cm}
\setlength{\belowcaptionskip}{0.3cm} 
	\begin{center}
		\includegraphics[width=0.85\linewidth]{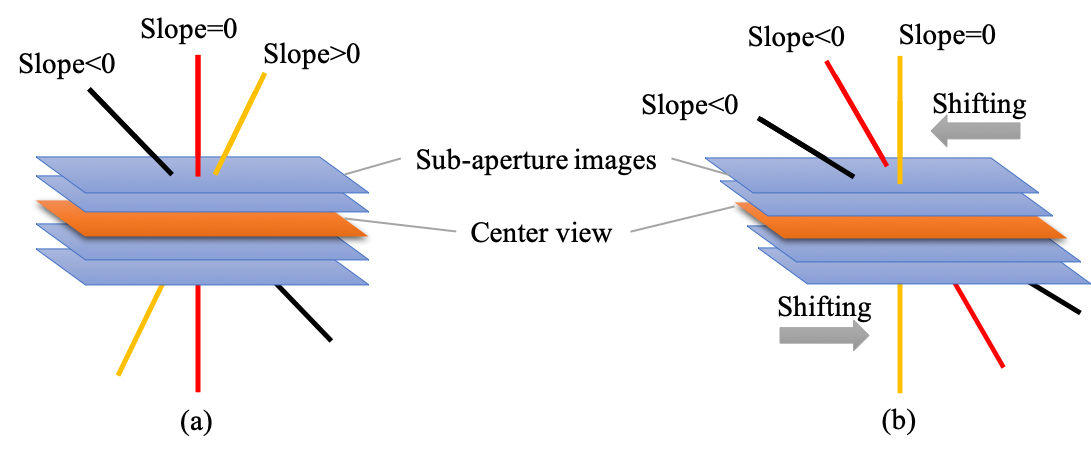}
	\end{center}
	\caption{The light field refocusing. (a) before refocusing. (b) after refocusing.}
	\label{fig:refocusing}
\end{figure}

To alleviate the problems of insufficient data and overfitting, we propose a new data augmentation method by refocusing EPIs. 
Differing from general augmentation techniques such as rotation, scaling and flipping~\cite{EPINET}, we refocus EPIs to generate multiple EPIs focused at different depth levels.
The light field refocusing shifts the sub-aperture images to obtain images focused at different depth planes~\cite{Goldluecke2013Epipolar}. 
Figure \ref{fig:refocusing} shows sub-aperture images at the same horizontal or vertical views that are stacked together. Lines with different slopes (i.e. the lines in EPIs) are inserted into the scene points of different depth planes on the sub-aperture images. The line at the focal depth should be vertical (slope = 0), while the other lines are inclined (slope $> 0$ or slope $< 0$). 
Taking the center view as the reference, the disparity shift between sub-aperture images changes the slope of the line. 
Thus, refocusing at a different depth plane changes the orientation of the structure in the EPI.

We convert the depth information into a disparity shift in every single EPI according to Eq. \ref{eq:depth}. 
The resulting disparity shift $\Delta x(u)$ related to the depth $Z$ is defined following \cite{Goldluecke2013Epipolar},
\begin{equation}
\Delta x(u) = (u_{0} - u)\frac{\Delta u}{Z} f
\label{eq:disparity shift}
\end{equation}
Here, we assume that the center view $(u_{0}, v_{0})$ is the reference view.
For the sake of simplicity, we also assume that lenslet-based cameras have the same focal length $f$ and the same baseline $\Delta u$ for the neighboring views.
Similarly, we can obtain the disparity shift $\Delta y(v)$. 
Then we refocus the EPI based on the refocusing principle \cite{Ren2006Digital},
\begin{equation}
L(u,v,x,y) = L(u,v,x + \Delta x(u),y + \Delta y(v))
\label{eq:refocusing EPI}
\end{equation}
The EPI patches $(a, b, c)$ in Figure \ref{fig:EPI} show three horizontal EPIs at different refocused depths.
Our strategy not only changes the slope of the orientation line but also changes the corresponding ground truth (Eq. \ref{eq:disparity shift}). 

\section{Experiments}

\subsection{Implementation Details}
Following previous works \cite{Luo,Feng2018}, we use the 4D light field benchmark \cite{HCIdataset2017} as our experimental dataset, which provides highly accurate disparity ground truth and performance evaluation metrics. 
The dataset includes $24$ carefully designed scenes with ground-truth disparity maps. 
Each scene has $9 \times 9$ angular resolution and $512 \times 512$ spatial resolution. 
$16$ scenes are used for training and the remaining $8$ scenes for testing. 
We randomly sample the horizontal and vertical EPI patch pairs of size $9 \times 29 \times 3$ from each scene as inputs. 
To avoid overfitting, we increase the training data to $8$ times the original data by the proposed EPI refocusing-based data augmentation. 

The bad pixel ratio (BadPix) \cite{HCIdataset2017}, which denotes the percentage of pixels whose disparity error is larger than 0.07 pixels, as well as the Mean Square Errors (MSE) are computed for evaluation metrics. 
Given an estimated disparity map $d$, the ground truth disparity map $gt$ and evaluation region $M$, BadPix is defined as,
\begin{equation}
{\rm{BadPix = }}\frac{{\left| {\left\{ {x \in M:\left| {d(x) - gt(x)} \right| > 0.07} \right\}} \right|}}{{\left| M \right|}},
\label{eq:BadPix}
\end{equation} 
and MSE is defined as,
\begin{equation}
{\rm{MSE}} = \frac{{\sum\limits_{x \in M} {{{\left( {d(x) - gt(x)} \right)}^2}} }}{{\left| M \right|}} \times 100.
\label{eq:MSE}
\end{equation}
Lower scores are better for both metrics. 

We use the Keras library \cite{keras2017} with the mean absolute error (MAE) loss to train the proposed network from scratch. 
We formularize the depth estimation as a multi-label regression problem to estimate the depth value of a single pixel.
Note that the network is trained end-to-end and does not make use of pre- and post-processing complications. We utilize the RMSprop optimizer \cite{RMSProp} and set the weight decay rate to $1e-5$ and batch size to $128$.
Our network training takes one day for $750k$ iterations on an NVIDIA GTX 1080 Ti 11GB GPU. 
The memory footprint is about $65\%$.

\subsection{Ablation Study}
We use the proposed network without the oriented relation module (ORM) and data augmentation based on EPI refocusing (EPIR) as the {\bf Baseline}.

\paragraph{Effect of the oriented relation module.}
Table \ref{table:Ablation} shows that the network using the ORM brings a significant improvement over the baseline, which can reduce the BadPix by around $1.5$.
\begin{table}[t!]
\setlength{\abovecaptionskip}{-0.3cm}
\setlength{\belowcaptionskip}{-0.1cm}
\footnotesize
	\begin{center}
		\begin{tabular}{|l|cccc|}
			\hline
			Metric & Baseline & w/ ORM & w/ EPIR & Full model\\
			\hline\hline
			BadPix & 9.45 & 7.98 & 7.41 & {\bf 5.66}\\
			MSE & 2.058 & 1.621 & 1.475 & {\bf 1.393}\\
			\hline
		\end{tabular}
	\end{center}
	\caption{Effects of ORM and EPIR. Bold: the best.}
	\label{table:Ablation}
\end{table}
Figure \ref{fig:Ablation} shows qualitative results for comparison.
\begin{figure}[t!]
\setlength{\abovecaptionskip}{-0.5cm}
\setlength{\belowcaptionskip}{0.4cm}
	\begin{center}
		\includegraphics[width=1\linewidth]{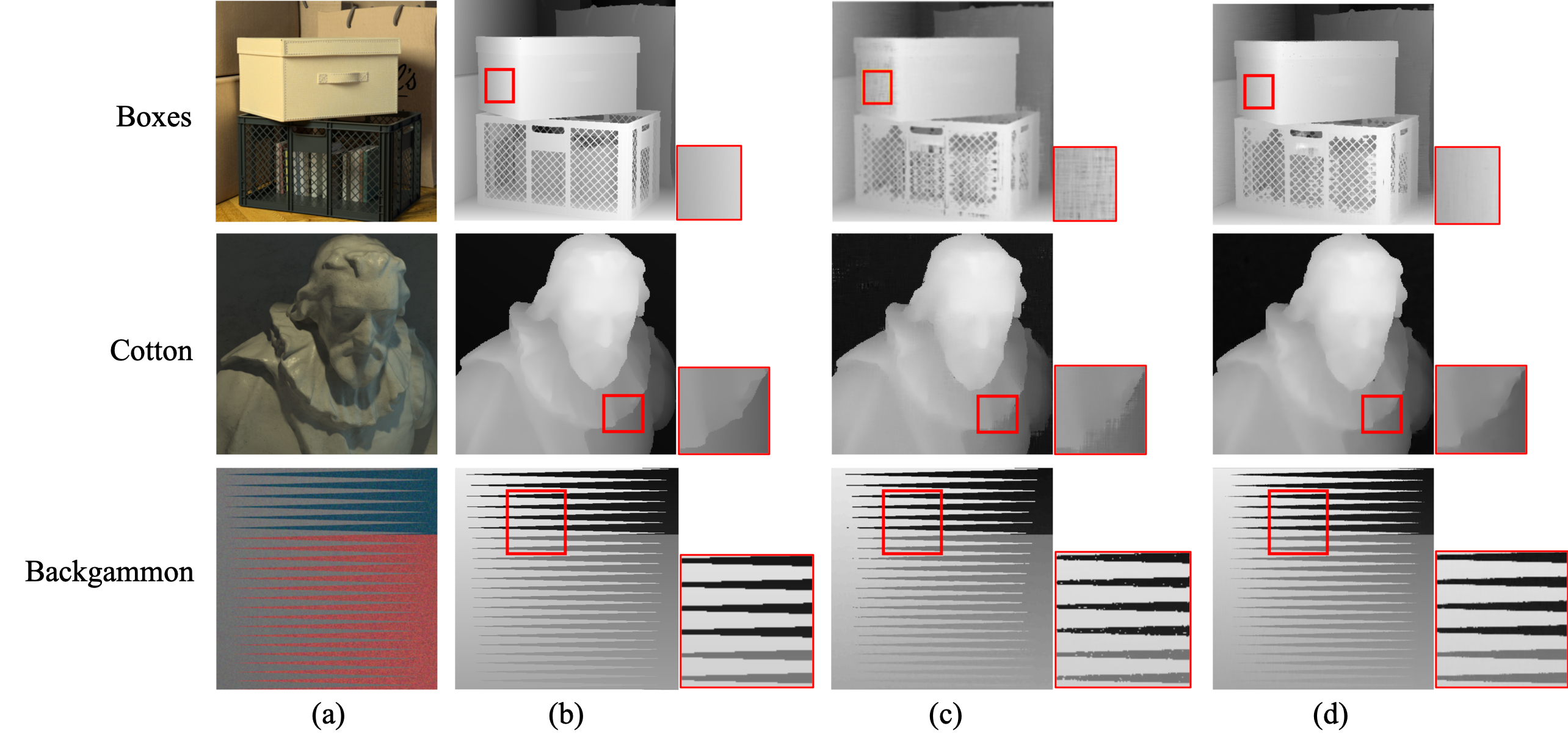}
	\end{center}
	\caption{Qualitative comparison of the baseline and the network with the ORM. (a) Original scenes. (b) Ground truth maps. (c) Baseline. (d) Our network with the ORM.}
	\label{fig:Ablation}
\end{figure}
{\em Boxes} and {\em Cotton} show that the ORM can reduce the streaking artifacts and improve the accuracy in weakly textured areas. 
The occlusion boundaries in {\em Backgammon} with multiple occlusions can also be better restored through the ORM.
In addition, our network with ORM generates smooth depth maps while preserving discontinuity between different objects, yielding the increased MSE by about $21\%$ compared to the baseline.

\vspace{-5 pt}
\paragraph{Effect of EPI refocusing-based data augmentation.}
From Table \ref{table:Ablation}, we can see that the network using the EPIR is better than the baseline.
Moreover, by using both the ORM and the EPIR, the performance is further boosted.
To further show the effect of EPI refocusing in the network, we compare the performance by varying the number of refocusing in Table \ref{table:augmentation}. 
We refocus the training data to the foreground and the background of the original depth plane.
From the table, we observe that there are performance gains when increasing the number of refocusing. 
However, there is no gain from EPIR$\times 8$ to EPIR$\times 10$ when comparing with Table \ref{table:augmentation}.
\begin{table}[t!]
\setlength{\abovecaptionskip}{-0.3cm}
\setlength{\belowcaptionskip}{0.4cm}
\footnotesize
	\begin{center}
		\begin{tabular}{|l|cccccc|}
			\hline
			Metric & Baseline & EPIR$\times 2$ & EPIR$\times 4$ & EPIR$\times 6$ & EPIR$\times 8$ & EPIR$\times 10$\\
			\hline\hline
			BadPix & 9.45 & 8.03 & 7.78 & 7.50 & {\bf 7.41} & 7.45 \\
			MSE & 2.058 & 1.781 & 1.511 & 1.482 & {\bf 1.475} & 1.480 \\
			\hline
		\end{tabular}
	\end{center}
	\caption{Performance in terms of the number of EPI refocusing. Bold: the best.}
	\label{table:augmentation}
\end{table}

\subsection{Comparison with State-of-the-Arts}
\begin{figure}[t!]
\setlength{\abovecaptionskip}{-0.5cm}
\setlength{\belowcaptionskip}{0.5cm} 
	\begin{center}
	\includegraphics[width=1\linewidth]{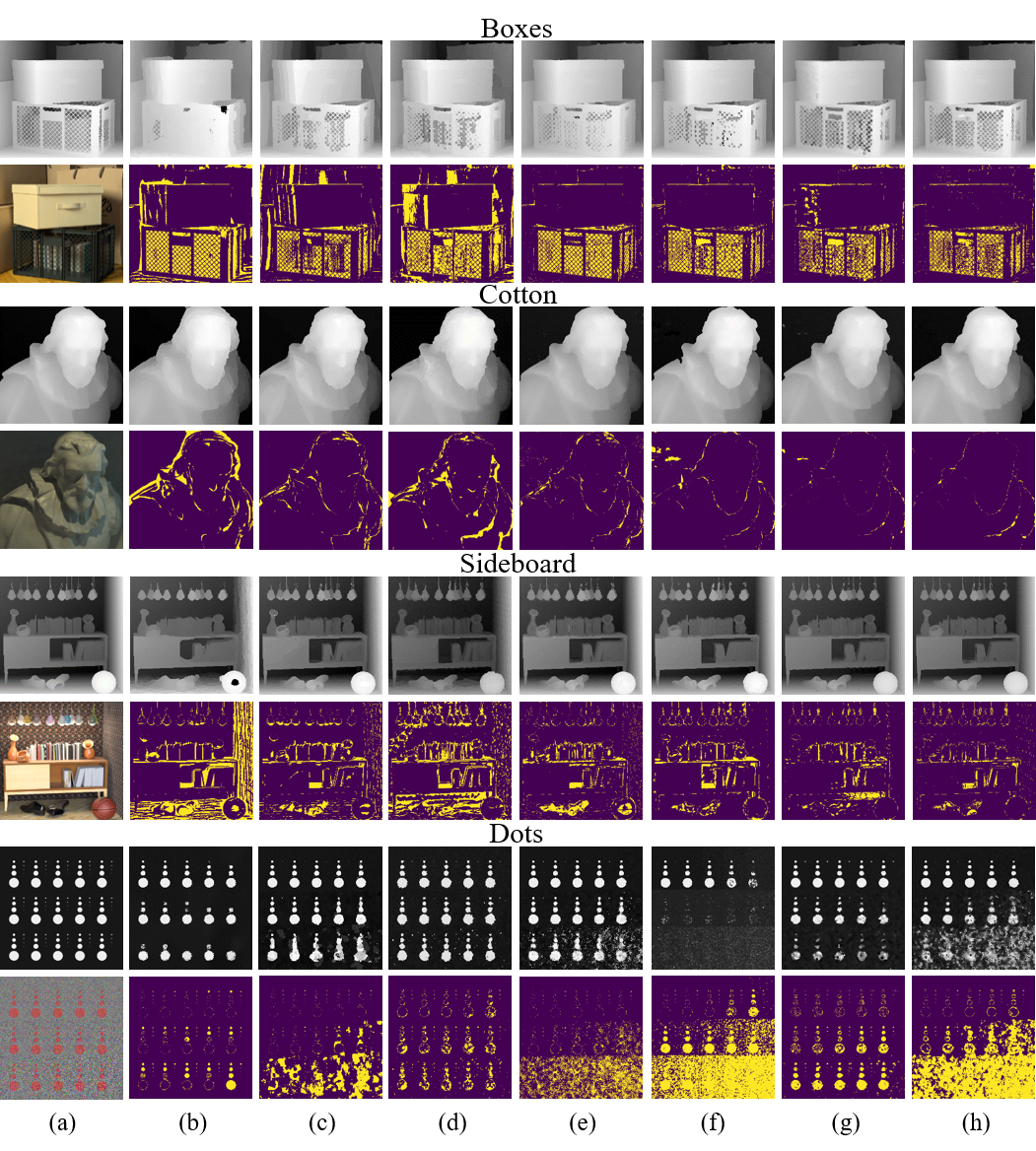}
	\end{center}
	\caption{Qualitative results on the 4D light field benchmark \cite{HCIdataset2017}. For each scene, the top row shows the estimated disparity maps and the bottom row shows the error maps for BadPix. (a) Ground truth. (b) LF~\cite{Jeon2015}. (c) CAE~\cite{Williem2016}. (d) LF\_OCC~\cite{Wang2015}. (e) SPO~\cite{ZhangSPO}. (f) EPN~\cite{Luo}. (g) EPINET~\cite{EPINET}. (h) Ours.}
	\label{fig:comparison}
\end{figure}

We compare our approach with other state-of-the-art methods: LF~\cite{Jeon2015}, CAE~\cite{Williem2016}, LF\_OCC ~\cite{Wang2015}, SPO~\cite{ZhangSPO}, EPN~\cite{Luo}, and EPINET~\cite{EPINET}.
The qualitative comparison is shown in Figure \ref{fig:comparison}.
The {\em Cotton} scene contains smooth surfaces and textureless regions, and the {\em Boxes} scene consists of occlusions with depth discontinuity. 
As can be seen from the figure, our approach can reconstruct the smooth surface and the region with sharp depth discontinuity compared to other methods. 
For the {\em Sideboard} scene with the complex shape and texture, our approach preserves more details and sharper boundaries by distinguishing the subtle difference of EPI slopes. 
In addition, our approach obtains better disparity maps in the {\em Boxes} and {\em Sideboard} scenes than the recent state-of-the-art method \cite{EPINET}, which uses the vertical, the horizontal, the left diagonal and the right diagonal viewpoints as inputs.
The number of the viewpoints is almost double that of our approach.
Compared with the 28-layer network of 4 branches in \cite{EPINET}, our network consists of 30 layers with 2 branches, which makes our trainable parameters be about half of those in \cite{EPINET}.
However, our network cannot produce good depth predictions for the {\em Dots} scene that contains a lot of noise, which is also the common downside of applying EPIs to the CNN-based method ({\em e.g.} EPN~\cite{Luo}).
The reason is that noises may lead to the false straight line estimation of EPI patches. Therefore, one of the future works could introduce global constraints of oriented lines into our model.

Quantitative results are shown in Tables \ref{table:BadPix} and \ref{table:MSE}, which show that the proposed approach performs the best in $4$ out of $8$ scenes.
\begin{table}[t!]
\setlength{\abovecaptionskip}{-0.3cm}
\setlength{\belowcaptionskip}{-0.3cm} 
\tiny 
\begin{center}
	\resizebox{\textwidth}{15mm}{
		\begin{tabular}{|l|ccccccc|}
			\hline
			Scenes & LF~\cite{Jeon2015} & CAE~\cite{Williem2016} &	LF\_OCC~\cite{Wang2015} & SPO~\cite{ZhangSPO} &	EPN~\cite{Luo} &	EPINET~\cite{EPINET} &	Ours \\
			\hline\hline
			boxes & 24.572 & 17.885 & 24.526 & 15.889 & \textcolor{green}{15.304} & \textcolor{blue}{14.190} & \textcolor{red}{13.373}\\
			cotton & 8.794 & 3.369 & 6.548 & 2.594 & \textcolor{green}{2.060} & \textcolor{red}{0.810} & \textcolor{blue}{0.869}\\
			dino & 21.478	& 4.968 & 15.466 & \textcolor{red}{2.184} & \textcolor{green}{2.877} & 2.970 & \textcolor{blue}{2.814}\\
			sideboard	& 23.906	& 9.845	& 17.923 & 9.297 & \textcolor{green}{7.997} & \textcolor{blue}{6.260} & \textcolor{red}{5.580}\\
			backgammon	& 4.810 & 3.924 & 18.061 & \textcolor{green}{3.781} & \textcolor{blue}{3.328}	& 4.130	& \textcolor{red}{2.511}\\
			dots	& \textcolor{red}{2.441} & 12.401 & \textcolor{blue}{5.109} & 16.274	& 39.248	& \textcolor{green}{9.370}	& 25.930\\
			pyramids	& 10.949 & 1.681	& 2.830	& \textcolor{green}{0.356}	& \textcolor{blue}{0.242}	& 0.540	& \textcolor{red}{0.240}\\
			stripes	& 35.394	& \textcolor{green}{7.872}	& 17.558 & 14.987 & 18.545	& \textcolor{red}{5.310}	& \textcolor{blue}{5.893}\\
			\hline
		 \end{tabular}}
		\end{center}
	\caption{Quantitative comparison of different methods using the BadPix metric. The best three results are shown in red, blue, and green, respectively (Best viewed in color).}
	\label{table:BadPix}
\end{table}
\begin{table}[t!]
\setlength{\abovecaptionskip}{-0.3cm}
\setlength{\belowcaptionskip}{0.3cm} 
\tiny 
\begin{center}
\resizebox{\textwidth}{15mm}{
		\begin{tabular}{|l|ccccccc|}
			\hline
			Scenes & LF~\cite{Jeon2015} & CAE~\cite{Williem2016} &	LF\_OCC~\cite{Wang2015} & SPO~\cite{ZhangSPO} &	EPN~\cite{Luo} &	EPINET~\cite{EPINET} &	Ours \\			\hline\hline
			boxes & 16.705 & \textcolor{green}{8.424} & 9.095 & 9.107 & 9.314 & \textcolor{blue}{6.440}	& \textcolor{red}{4.189}\\
			cotton &11.773	& 1.506	& \textcolor{green}{1.103}	& 1.313	& 1.406	& \textcolor{red}{0.270}	& \textcolor{blue}{0.313}\\
			dino & 1.558	& \textcolor{green}{0.382}	& 1.077	& \textcolor{red}{0.310}	& 0.565	& 0.940	& \textcolor{blue}{0.336}\\
			sideboard	& 4.735 & \textcolor{green}{0.876} & 2.158 & 1.024 & 1.744	& \textcolor{blue}{0.770}	& \textcolor{red}{0.733}\\
			backgammon	&15.109	& 6.074	& 20.962 & \textcolor{green}{4.587} & \textcolor{blue}{3.699} & 4.700 & \textcolor{red}{1.403}\\
			dots & \textcolor{green}{4.803}	& 5.082	& \textcolor{red}{2.731}	& 5.238	& 22.369	& \textcolor{blue}{3.320}	& 6.754\\
			pyramids & 0.243	& 0.048	& 0.098	& 0.043	&\textcolor{blue}{0.018} 	& \textcolor{green}{0.020}	& \textcolor{red}{0.016}\\
			stripes	& 17.380	& \textcolor{green}{3.556}	& 7.646	& 6.955	& 8.731	& \textcolor{red}{1.160}	& \textcolor{blue}{1.263}\\
			\hline
		\end{tabular}}
	  \end{center}
	\caption{Quantitative comparison of different methods using the MSE metric. The best three results are shown in red, blue, and green, respectively (Best viewed in color).}
	\label{table:MSE}
\end{table}
In particular, the proposed approach predicts more accurate disparity values on the {\em Boxes} and {\em Backgammon} scenes under multi-occlusions. 
Note that we do not apply any post-processing for depth optimization while most other methods \cite{Jeon2015,Williem2016,Wang2015,ZhangSPO,Luo} are accompanied by post optimization.

\section{Conclusion}
In this paper, we propose an end-to-end fully convolutional network for depth estimation from light fields by exploiting horizontal and vertical EPIs. 
We introduce a new relational reasoning module to construct the relationship between oriented lines in EPIs. 
In addition, we propose a new data augmentation method by refocusing the EPIs.
We demonstrate the effectiveness of our approach on the 4D light field benchmark \cite{HCIdataset2017}.
Our approach is competitive with the state-of-the-art methods, and is able to predict more accurate disparity map in some challenging scenes such as {\em Boxes} and {\em Sideboard} without any post-processing.

\vspace{-5 pt}
\paragraph{Acknowledgments.} This work was supported by the National Natural Science Foundation of China, No. 61876057.


\begin{thebibliography}{10}
	
	\bibitem{Goldluecke2013Epipolar}
	Maximilian. Diebold and Bastian. Goldluecke.
	\newblock Epipolar Plane Image Refocusing for Improved Depth Estimation and Occlusion Handling.
	\newblock In {\em Vision, Modelling \verb'&' Visualization},
	\newblock 2013.
	
	\bibitem{Feng2018}
	M. Feng, Y. Wang, J. Liu, L. Zhang, H. F. M. Zaki, and A. Mian.
	\newblock Benchmark Data Set and Method for Depth Estimation From Light Field Images.
	\newblock {\em IEEE Transactions on Image Processing},
	\newblock 2018.
	
	\bibitem{He2016}
	K. He, X. Zhang, S. Ren, and J. Sun.
	\newblock Deep Residual Learning for Image Recognition.
	\newblock In {\em Proceedings of IEEE Conference on Computer Vision and Pattern Recognition (CVPR)},
	\newblock 2016.
	
	\bibitem{HeberandPock2016}
	S. Heber and T. Pock.
	\newblock Convolutional Networks for Shape from Light Field.
	\newblock In {\em Proceedings of IEEE Conference on Computer Vision and Pattern Recognition (CVPR)},
	\newblock 2016.
	
	\bibitem{Heber2017}
	S. Heber, W. Yu, and T. Pock.
	\newblock Neural EPI-Volume Networks for Shape from Light Field.
	\newblock In {\em Proceedings of International Conference on Computer Vision (ICCV)},
	\newblock 2017.
	
	\bibitem{Heber2016}
	Stefan. Heber, Yu. Wei, and Thomas. Pock.
	\newblock U-shaped Networks for Shape from Light Field.
	\newblock In {\em Proceedings of British Machine Vision Conference},
	\newblock 2016.
	
	\bibitem{HCIdataset2017}
	K. Honauer, O. Johannsen, D. Kondermann, and B. Goldluecke.
	\newblock A Dataset and Evaluation Methodology for Depth Estimation on 4D Light Fields.
	\newblock In {\em Proceedings of Asian Conference on Computer Vision (ACCV)},
	\newblock 2016.
	
	\bibitem{Hu2018}
	H. Hu, J. Gu, Z. Zhang, J. Dai, and Y. Wei.
	\newblock Relation Networks for Object Detection.
	\newblock In {\em Proceedings of IEEE Conference on Computer Vision and Pattern Recognition (CVPR)},
	\newblock 2018.
	
	\bibitem{Jeon2015}
	H. Jeon, J. Park, G. Choe, J. Park, Y. Bok, Y. Tai, and I. S. Kweon.
	\newblock Accurate depth map estimation from a lenslet light field camera.
	\newblock In {\em Proceedings of IEEE Conference on Computer Vision and Pattern Recognition (CVPR)},
	\newblock 2015.
	
	\bibitem{keras2017}
	Z. Jiang and G. Shen.
	\newblock Prediction of House Price Based on The Back Propagation Neural Network in The Keras Deep Learning Framework.
	\newblock In {\em Proceedings of International Conference on Systems and Informatics (ICSAI)},
	\newblock 2019.
	
	\bibitem{op2013}
	Ole. Johannsen, Christian. Heinze, Bastian. Goldluecke, and Perwa. Christian.
	\newblock On the Calibration of Focused Plenoptic Cameras.
	\newblock Springer Berlin Heidelberg,
	\newblock 2013.
	
	\bibitem{EPI-shift}
	T. Leistner, H. Schilling, R. Mackowiak, S. Gumhold, and C. Rother.
	\newblock S. Gumhold, and C. Rother. Learning to think outside the box: Wide-baseline light field depth estimation with EPI-shift.
	\newblock In {\em Proceedings of 2019 International Conference on 3D Vision (3DV)},
	\newblock 2019.
	
	
	\bibitem{LFrendering}
	Marc. Levoy and Pat. Hanrahan.
	\newblock Light Field Rendering.
	\newblock In {\em Proceedings of the 23rd Annual Conference on Computer Graphics and Interactive Techniques},
	\newblock 1996.
	
	\bibitem{focalStack2015}
	H. Lin, C. Chen, S. B. Kang, and J. Yu.
	\newblock Depth Recovery from Light Field Using Focal Stack Symmetry.
	\newblock In {\em Proceedings of International Conference on Computer Vision (ICCV)},
	\newblock 2015.
	
	\bibitem{Luo}
	Y. Luo, W. Zhou, J. Fang, L. Liang, H. Zhang, and G. Dai.
	\newblock EPI-Patch Based Convolutional Neural Network for Depth Estimation on 4D Light Field.
	\newblock In {\em International Conference on Neural Information Processing},
	\newblock 2017.
	
	\bibitem{Mou2019}
	L. Mou, Y. Hua, and X. X. Zhu.
	\newblock A Relation-Augmented Fully Convolutional Network for Semantic Segmentation in Aerial Scenes.
	\newblock In {\em Proceedings of IEEE Conference on Computer Vision and Pattern Recognition (CVPR)},
	\newblock 2019.
	
	
	\bibitem{Hanrahan2005Light}
    Ren. Ng, Marc. Levoy, Gene. Duval, Mark. Horowitz, and Pat. Hanrahan.
	\newblock Light Field Photography with a Hand-held Plenoptic Camera.
	\newblock {\em Computer Science Technical Report CSTR}, 2005. 
	
	
	\bibitem{Ren2006Digital}
	Ren. Ng.
	\newblock Digital light field photography.
	\newblock PhD thesis, Stanford University,
	\newblock 2006.
	
	\bibitem{EPINET}
	C. Shin, H. Jeon, Y. Yoon, I. S. Kweon, and S. J. Kim.
	\newblock EPINET: A Fully-Convolutional Neural Network Using Epipolar Geometry for Depth from Light Field Images.
	\newblock In {\em Proceedings of IEEE Conference on Computer Vision and Pattern Recognition (CVPR)},
	\newblock 2018.
	
	
	\bibitem{Tao2013}
	M. W. Tao, S. Hadap, J. Malik, and R. Ramamoorthi.
	\newblock Depth from Combining Defocus and Correspondence Using Light-Field Cameras.
	\newblock In {\em Proceedings of International Conference on Computer Vision (ICCV)},
	\newblock 2013.
	
	\bibitem{Wang2015}
	T.-C. Wang, A. Efros, and R. Ramamoorthi.
	\newblock Occlusion-Aware Depth Estimation Using Light-Field Cameras.
	\newblock In {\em Proceedings of International Conference on Computer Vision (ICCV)},
	\newblock 2017.
	
	
	\bibitem{Wanner2012Globally}
	S. Wanner and B. Goldluecke.
	\newblock Globally Consistent Depth Labeling of 4D Light Fields.
	\newblock In {\em Proceedings of IEEE Conference on Computer Vision and Pattern Recognition (CVPR)},
	\newblock 2012.
	
	\bibitem{Wanner2014}
	S. Wanner and B. Goldluecke.
	\newblock Variational Light Field Analysis for Disparity Estimation and Super-Resolution.
	\newblock {\em IEEE Transactions on Pattern Analysis and Machine Intelligence (TPAMI)},
	\newblock 2014.
	
	\bibitem{Williem2016}
	W. Williem and I. K. Park.
	\newblock Robust Light Field Depth Estimation for Noisy Scene with Occlusion.
	\newblock In {\em Proceedings of IEEE Conference on Computer Vision and Pattern Recognition (CVPR)},
	\newblock 2016.
	
	\bibitem{Sun2016}
	Xing. Sun, Z. Xu, Nan. Meng, E. Y. Lam, and H. K. -H. So.
	\newblock Data-driven light field depth estimation using deep Convolutional Neural Networks.
	\newblock In {\em Proceedings of International Joint Conference on Neural Networks (IJCNN)},
	\newblock 2016.
	
	\bibitem{siamese2015}
	S. Zagoruyko and N. Komodakis.
	\newblock Learning to compare image patches via convolutional neural networks.
	\newblock In {\em Proceedings of IEEE Conference on Computer Vision and Pattern Recognition (CVPR)},
	\newblock 2015.
	
	\bibitem{ZhangSPO}
	S. Zhang, H. Sheng, C. Li, J. Zhang, and Z. Xiong.
	\newblock Robust depth estimation for light field via spinning parallelogram operator.
	\newblock {\em Computer Vision and Image Understanding (CVIU)},
	\newblock 2016.
	
	\bibitem{YZhang}
	Y. Zhang, H. Lv, Y. Liu, H. Wang, X. Wang, Q. Huang, X. Xiang, and Q. Dai.
	\newblock Light-Field Depth Estimation via Epipolar Plane Image Analysis and Locally Linear Embedding.
	\newblock {\em IEEE Transactions on Circuits and Systems for Video Technology},
	\newblock 2017.
	
	\bibitem{Zhou2018}
	B. Zhou, A. Andonian, and A. Torralba.
	\newblock Temporal Relational Reasoning in Videos.
	\newblock In {\em Proceedings of European Conference on Computer Vision (ECCV)},
	\newblock 2018.
	
	
	\bibitem{RMSProp}
	F. Zou, L. Shen, Z. Jie, W. Zhang, and W. Liu.
	\newblock A Sufficient Condition for Convergences of Adam and RMSProp.
	\newblock In {\em Proceedings of IEEE Conference on Computer Vision and Pattern Recognition (CVPR)},
	\newblock 2019.
	


	
	
	
	
	


	
\end{thebibliography}

\end{document}